\documentclass{article}

\usepackage{arxiv}

\usepackage[utf8]{inputenc} 
\usepackage[T1]{fontenc}    
\usepackage{hyperref}       
\usepackage{url}            
\usepackage{booktabs}       
\usepackage{amsfonts}       
\usepackage{nicefrac}       
\usepackage{microtype}      
\usepackage{lipsum}
\usepackage{cite}
\usepackage[pdftex]{graphicx} 
\usepackage{amsmath}
\usepackage{graphicx}
\usepackage[section]{placeins}
\usepackage{multirow}
\usepackage{multicol}

\title{Estimating the Brittleness of AI: Safety Integrity Levels and the Need for Testing Out-Of-Distribution Performance}

\author{
  Andrew J. Lohn\\
  Georgetown University, Walsh School of Foreign Service\\
  Washington, DC, USA \\
  \texttt{drew.lohn@georgetown.edu} \\
}

\begin{document}
\maketitle

\begin{abstract}
Test, Evaluation, Verification, and Validation (TEVV) for Artificial Intelligence (AI) is a challenge that threatens to limit the economic and societal rewards that AI researchers have devoted themselves to producing. A central task of TEVV for AI is estimating brittleness, where brittleness implies that the system functions well within some bounds and poorly outside of those bounds. This paper argues that neither of those criteria are certain of Deep Neural Networks. First, highly touted AI successes (eg. image classification and speech recognition) are orders of magnitude more failure-prone than are typically certified in critical systems even within design bounds (perfectly in-distribution sampling). Second, performance falls off only gradually as inputs become further Out-Of-Distribution (OOD). Enhanced emphasis is needed on designing systems that are resilient despite failure-prone AI components as well as on evaluating and improving OOD performance in order to get AI to where it can clear the challenging hurdles of TEVV and certification.
\end{abstract}


\section{Introduction}
\label{sec:introduction}
Test, Evaluation, Verification, and Validation (TEVV) for Artificial Intelligence (AI) is a central challenge that threatens to limit the economic and societal rewards that AI researchers have devoted themselves to producing.\cite{afrlLicense,dibEthics,jaicUnderstandingAI,tarraf,concreteProblems} TEVV for AI is particularly challenging for several reasons\cite{idaTest,borg} including that AI is meant to be used in circumstances that the designers cannot fully envision. The decision about what actions to take is made by the AI during use rather than by the designer before testing. The benefit is that during use, the AI will have access to the full range of inputs and environmental data to make decisions. The drawback is that it is too late for additional TEVV at that point.

For example, non-AI software for autonomous braking would have well-delineated responses to the sensors such as radar and velocity. The decisions about how the vehicle will respond are made by the designer at design time and written into software which can have its code reviewed and tested to ensure that responses match the design decisions. For an AI, the designer has not made explicit decisions about how the vehicle will behave. Those decisions are made at use time, after testing and certification has already been completed. 

In cases where the range of inputs and environments that can be encountered are well understood, then testing is a straightforward matter. It may be arduous and expensive to gather enough test cases to certify the system, but it is conceptually simple to do. What makes TEVV for AI difficult is that often the range of inputs and environments that can be encountered is not well understood. The system is likely to encounter situations that are outside of the distributions of scenarios it was designed for.

AI has been known to fail in those circumstances and has been commonly disparaged as brittle.\cite{cummingsBrittle,heavenBrittle,leetaruBrittle,pontinBrittle} Brittleness implies two things about a component, first that it is highly functioning within some bounds and second, that it breaks readily when those bounds are exceeded. This report argues that neither of those criteria are as plain as often presented. First, compared to the reliability required of safety- or mission-critical systems for which TEVV and certification are paramount, the most highly touted AI successes are orders of magnitude more failure-prone even when being evaluated on data drawn from the same distributions they were designed for. And second, the performance of those models degrades smoothly as those bounds on the data distributions are relaxed, at a rate that is sometimes comparable to humans.

This report starts by briefly familiarizing the reader with the existing approaches for certifying critical systems in section 2 then expands on the required levels of reliability for components in critical systems in section 3. Section 4 contextualizes those required reliability levels for AI. Section 5 presents experimental results evaluating the brittleness of some AI algorithms and section 6 concludes.

\section{Certification in Critical Systems}
\label{sec:criticalsystems}
\subsection{Processes and Standards are Complex and Opaque}
AI is a ubiquitous technology that can be envisioned in an infinity of applications. Many of those applications warrant deeper levels of scrutiny than others because of heightened risks and, for the most part, those applications have already evolved processes for obtaining high levels of surety. Those processes tend to be complex and opaque to an extent where people build entire careers around understanding and navigating them. Additionally, most of the standards documents (the notable exception being military standards) are behind paywalls, restricting access and widespread understanding. As a result of the opacity, complexity, and diversity of these processes, it is probably not feasible for most AI designers to be well-versed in even one of them let alone the wide range of applications for which a single model or architecture may be used.

Nevertheless, currently much of the onus for AI safety, and much of the blame for failures, falls on the AI designers. While recognizing that the enormity of the problem will require many dedicated staff whose sole job is translating safety goals into design specifications, all involved should have a basic understanding of the processes, requirements, and goals. The next few subsections will very briefly highlight a few of the takeaways that are relevant to AI designers and practitioners from a survey of the processes for certifying aircraft, nuclear power plants, automobiles, pharmaceuticals, and weapons systems. Their processes and standards are far from perfect as is recognized by the certifying bodies themselves who have sponsored studies\cite{faaAlternatives,nhstaAlternatives} to search out better alternatives. There have been major and minor revisions to the standards but the existing approaches seem to hold up well to scrutiny as compared to other options that have been proposed.

\subsection{Is AI Software or an Operator}
The certification of new critical systems (such as a new aircraft or a power plant design) moves along established processes and standards which are each composed of many lower level standards and processes. The chains can be extensive. Among those lower level standards can be those dedicated to software components and those dedicated to operators. Although AI is clearly software (or perhaps electronic hardware), depending on the intended application it might be used to perform tasks that are more commonly associated with operators. Certification of operators tends to rely on passing tests, accumulating hours of experience, and continually monitoring fitness for duty. A driver's license is a familiar example that requires a written and practical exam, starts with a permit, and can be revoked for poor performance or degraded operational capacity. More stringent examples are pilots\cite{pilotRegs} and nuclear operators\cite{nureg1021} which require minimum education and minimum hours of supervised operation.

The processes and standards that exist for operators were designed for humans and are not easily applied to AI so even in cases where the tasks being performed are more commonly associated with those of an operator than of traditional software, the processes and standards for software certification tend to be more appropriate for AI. Testing and certification that licenses AI as more of an operator is a promising direction of study\cite{afrlLicense} but does not appear to be a near-term solution. An important distinction to consider in operator-like approaches vs software-like approaches is that the diversity of human operators provides some assurance against systemic failures across the entire fleet but also precludes in-depth testing of the entire set of operators in a cost and time-efficient manner. That systemic risk and opportunity for in-depth testing is a main reason why software standards and processes are currently more applicable to AI.

\subsection{Various Approaches to Providing Surety}
At a high level, there are many different approaches that can be used to provide increased levels of surety. Below is a partial list illustrating some of them:
\begin{itemize}
\item Find and fix faults before fielding
\item Fail a large fraction of prospect products
\item Detect and respond to failures in real time
\item Approve limited quantities or functionalities and observe real-world performance
\item Penalize organizations that provide faulty products
\end{itemize}

In actual practice, all critical application areas implement all of these surety approaches and more. They are each applied to varying degrees with some approaches more or less prevalent in some applications than in others. Autonomous vehicles have famously\cite{kalra} approved limited quantities of vehicles in confined locations for evaluation. The same approach is less useful for nuclear power plants where the scale of catastrophe from a single complete failure could have a much larger impact that is less confined to a location. Nuclear power instead follows a probabilistic risk assessment method\cite{keller} to try to minimize the likelihood of failures before operation starts. The U.S. military's conventional weapons systems face a scale of risk that is intermediate between car accidents and radiological disaster so put an emphasis on finding and fixing faults before fielding but systems also then step through Initial Operating Capability before proceeding to Full Deployment.\cite{dodi5000} New pharmaceuticals also carry potentially wide-spread risks but have adopted a somewhat different strategy that produces many candidates, most of which ultimately fail the certification process rather than fixing flaws in the products.\cite{vanNorman} 

Some approaches to surety have little to do with testing or certification directly. For example, automotive companies have been charged with the expense of liability or forced to recall defective products when safety issues are discovered.\cite{gurney} Similarly, medical doctors are tested and certified but then are also held accountable for malpractice, how this will play out for AI and robotics as they take on a larger role in society is still being decided.\cite{chung,datteri}

\section{Safety Integrity Levels}
\label{sec:sils}
Typical among many of the safety standards for critcal systems is the concept of levels of safety, somewhat akin to tolerable failure rates. Industries often have their own set of levels that are different in magnitudes, labeling, intent, and implementation. To think of them as simply tolerable failure rates is admittedly a disservice to the standards and to safety culture. Some standards distance themselves from quantitative failure rates, especially for software such as MIL-STD-882E\cite{milStandard882E} which states that: "determining the probability of failure of a single software function is difficult at best and cannot be based on historical data," or DO-178C\cite{do178C} which states: "it is important to realize that the likelihood that software contains an error cannot be quantified in the same way as for random hardware failures." Nonetheless, it is necessary for AI designers to have a rough guidepost from which to understand the order of magnitude of the surety problem they face and the reliability requirements of the systems they are designing. With that goal in mind, a few of the standards and their target failure rates will be listed here along with a brief summary of how they are used.

\subsection{Various SILs}
Aircraft use Design Assurance Level (DAL) from DO-178C: "Software Considerations in Airborne Systems and Equipment Certification." Aviation also separately uses numbered Software Assurance Levels (AL) from DO-278: "Guidelines For Communication, Navigation, Surveillance, and Air Traffic Management (CNS/ATM)."\cite{faaGround}\cite{brosgol} The automotive industry uses Automotive Safety Integrity Level (ASIL) from ISO 26262: "Road vehicles - Functional Safety." \cite{nardi} A common industry-agnostic standard is provided in IEC 61508: "Functional Safety of Electrical/Electronic/Programmable Electronic Safety-related Systems," which uses the generic term Safety Integrity Level (SIL).\cite{rausand} The level of failure per hour and corresponding labels for each industry standard are provided in Table \ref{tab:SILfailrates}

\begin{table}[!htb]
    \centering
    \begin{tabular}{|c|c|c|c|c|c|c|c|c|}
          \hline
           & $10^{-9}/h$ & $10^{-8}/h$ & $10^{-7}/h$ & $10^{-6}/h$ & $10^{-5}/h$ & $10^{-4}/h$ & $10^{-3}/h$ & $>10^{-3}/h$ \\ \hline
         Automotive &  & D & C/B & A &  &  &  &  \\ \hline
         Aviation & A &  & B & & C &  & D & E  \\ \hline
         CNS/ATM & AL1 &  & AL2 &  & AL3 & AL4 & AL5 & AL6 \\ \hline
         IEC 61508 &  & 4 & 3 & 2 & 1 &  &  &  \\ \hline
    \end{tabular}
    \caption{Approximate hourly failure rates are provided for the range of levels of surety in various industries.}
    \label{tab:SILfailrates}
\end{table}

Aviation standards cover a wide range from one in a thousand hours for low risk systems to one in a billion hours for more critical systems. Automotive standards in contrast have a slightly lower requirement for their most critical systems but only begin to have specified standards at the one in a million hours level.

The industry-agnostic IEC 61508 standard offers a second unit of measurement that can be applied when hourly rates are not appropriate. A separation of SILs into Continuous Demand vs Low Demand requirements allows for systems that are used less frequently to be evaluated on a per use basis rather than a per hour basis. The Low Demand SILs are shown in Table \ref{tab:lowdemandSIL}.\cite{rausand}

\begin{table}[!htb]
    \centering
    \begin{tabular}{|c|c|c|c|c|}
        \hline
         & $10^{-4}/use$ & $10^{-3}/use$ & $10^{-2}/use$ & $10^{-1}/use$ \\ \hline
         IEC 61508 & 4 & 3 & 2 & 1 \\ \hline
    \end{tabular}
    \caption{Low demand safety levels are evaluated on a per use basis rather than a per hour basis.}
    \label{tab:lowdemandSIL}
\end{table}

\subsection{Examples of SILs}
Although the failure rates for the SILs is quantitative, deciding which level of assurance should be applied to a given system is not as easily quantified. Industries vary in how this task is performed as well. Aircraft for example have categories of Catastrophic, Hazardous, Major, Minor, and No Effect corresponding to their DALs and use qualitative and quantitative definitions of likelihoods. MIL-STD-882, "System Safety," uses qualitative severity and probability levels with a mix of qualitative and quantitative elements in their definitions. The automotive industry (ISO 26262) breaks risk into three factors: Severity, Likelihood (called "Exposure") and Controllability, each of which has several qualitative levels as described in the list below:
\begin{itemize}
    \item Severity
    \begin{itemize}
        \item S1: Light and moderate injuries
        \item S2: Severe and life-threatening injuries (survival probable)
        \item S3: Life-threatening injuries (survival uncertain)
    \end{itemize}
    \item Exposure
        \begin{itemize}
        \item E1: Very low probability
        \item E2: Low probability
        \item E3: Medium probability
        \item E4: High probability
    \end{itemize}
    \item Controllability
        \begin{itemize}
        \item C1: Simply controllable
        \item C2: Normally controllable
        \item C3: Difficult to control or uncontrollable
    \end{itemize}
\end{itemize}

Combining a qualitative estimate for each of Severity, Exposure, and Controllability gives a risk to which an ASIL can be assigned. All of the possible combinations of Severity, Exposure, and Controllability are shown in Table \ref{tab:qualitativeSIL} along with the assigned ASIL. Only the the most risky combination of the three merits an ASIL D, and most of the lowest risk combinations do not merit any ASIL at all.

\begin{table}[!htb]
    \centering
    \begin{tabular}{|c|c|c|c|c|}
        \cline{3-5}
         \multicolumn{2}{c|}{} & C1 & C2 & C3 \\ \hline
         \multirow{4}{*}{S1} & E1 & - & - & - \\ \cline{2-5}
          & E2 & - & - & - \\ \cline{2-5}
          & E3 & - & - & A \\ \cline{2-5}
          & E4 & - & A & B \\ \hline
         
          \multirow{4}{*}{S2} & E1 & - & - & - \\ \cline{2-5}
          & E2 & - & - & A \\ \cline{2-5}
          & E3 & - & A & B \\ \cline{2-5}
          & E4 & A & B & C \\ \hline
         
          \multirow{4}{*}{S3} & E1 & - & - & A \\ \cline{2-5}
          & E2 & - & A & B \\ \cline{2-5}
          & E3 & A & B & C \\ \cline{2-5}
          & E4 & B & C & D \\ \hline
    \end{tabular}
    \caption{The automotive industry is an example of using a qualitative approach to assigning safety levels to sub-systems.}
    \label{tab:qualitativeSIL}
\end{table}

To add intuition, a few examples of ASIL D (highest automotive level) are inadvertent airbag deployment and some types of unwanted deceleration, braking or acceleration or self-steering failures. Level C includes some types of unintended deceleration, braking, or acceleration. ASIL B includes outages in front or rear view cameras or brake lights. ASIL A includes failures of rear lights on both sides.\cite{ifineonAuto} With increasing amounts of automation, human drivers will be less available to help with controllability moving some of these systems to higher ASILs.

Aviation examples use the opposite lettering direction where A is the highest level and E is the lowest. A and B DALs include systems like fly-by-wire controls, auto-pilot, radar, Identification Friend or Foe (IFF), and missile launch. Example at the C and D levels include anti-missile defense, telemetry, weapons targeting.\cite{horvath}

\subsection{Achieving SIL with Less Reliable Components}
As is clear from the examples in the previous section, the SILs are applied to systems or functions at a higher abstraction than the individual components, so the components do not necessarily need to achieve the SIL's failure rates individually. It is possible to build a system that has higher reliability than its constituent parts. For example, allowing for proper independence and redundancy, it is possible to decompose ASIL-D into two ASIL-B components or to achieve ASIL-C with an ASIL-A and an ASIL-B component.\cite{alcaide} Instead, it may also be possible to increase the controllability by giving a human the opportunity to intervene. It may also be possible to  place limits on the operating environment to decrease the exposure (likelihood) such as by limiting autonomous vehicles to sunny dry geographies rather than snowy ones.

\section{SILs for AI}
\label{sec:silsforai}
With a rough intuition for the safety goals to achieve, one can begin to assess the feasibility of developing various AI technologies that reach the required reliability as well as developing the testing methods that would be needed to certify them.

\subsection{Reliability in AI}
AI is typically tasked with difficult problems. For example, performance on Image-Net classification has been one of AI's most lauded successes. At the time of writing, it is led by groups posting top-1 accuracies in the upper 80's of percent and top-5 accuracies in the upper 90's\cite{touvron}\cite{xieImageNet}\cite{kolesnikov} and disease detection from medical imaging is comparable.\cite{liuDiagnoses} That incredible level of performance is the result of the combined efforts of a massive number of researchers over many years but still the failure rates are greater than $10^{-2}/use$. Those failure rates correspond to the lowest SIL rating in all of Table \ref{tab:lowdemandSIL} for any industry. When considered in a continuous setting where many images are evaluated per hour, the situation is far worse. In that case, the failure rates are too high for any SIL. To reach even the minimum continuous-demand level would require more than ten hours between attempts for top-5 images and more than 100 hours between attempts for top-1 images. Those attempt frequencies are far from what most AI practitioners consider continuous. Processing even just ten images per second would require an accuracy of 0.99999997 to get to even the lowest level in aviation $10^{-3}/h$, and the lowest level for the automotive industry is three orders of magnitude more restrictive than that. Viewed in this light, AI is not confronted with a testing challenge, it is facing a reliability challenge.

Getting to one-in-a-million failure rates in image classification for diverse sets of objects is perhaps an unfair goal. Image-Net classifiers already outperform humans. That said, critical systems and their certification procedures have been designed over a century to accommodate humans as a fallible component of a reasonably reliable system. Further, as mentioned earlier, certification of AI as an operator rather than as software does not appear to be a near-term path forward. Either way, the message to be received by AI designers and practitioners is that the failure rates of the most heralded algorithms are orders of magnitude more failure-prone than safety-critical systems typically certify.

\subsection{The Challenge of OOD}
More importantly than getting to one-in-a-million failure rates is getting to one-in-a-million failure rates in the operational environment. The standard approach for evaluating an AI model's reliability is to measure performance in a hold-out dataset that the model has never been exposed to before. But that dataset is almost always drawn randomly from the same set of samples as those the model has been trained to. In real-world operations it is likely that some or many of the situations the model will encounter will be different in some way from the types of samples used in training and testing. These situations are called Out-Of-Distribution (OOD) and the study of OOD in machine learning is currently gaining popularity.

\subsection{OOD and Domain Adaptation}
Much of the focus within the literature has been on techniques for OOD detection and for Domain Adaptation (DA). OOD detection is the task of trying to determine by various means whether an input is OOD. There are many ways that this is done and new methods are being discovered or invented constantly. Some methods compare the new inputs to the set of inputs in the training set, some methods assess the relative confidence across the outputs of the model, other methods evaluate internal parameters of the model such as gradients.\cite{bulusu}

Domain Adaptation \cite{daume} is a different class of techniques that are meant to increase the effectiveness of a model on inputs that are from a different distribution than those it was originally trained on. DA is typically useful when a large data source is available but when that data source is not perfectly matched to the desired problem. If that data source has the same outputs as the intended problem but different inputs then DA is applied. If the inputs from the two sources are from the same distribution but the outputs are not then the process is called transfer learning. An example of DA is the use of digitally-generated video game footage as a training substitute for real-world training samples. The input distribution in the intended application (real-world) is of a different type than in the large body of training data (video game footage) but the intended output task is the same for both training and application (i.e identify pedestrians).

Both of these streams of research are making valuable contributions to increasing the reliability of AI in critical applications where real-world applications are slightly different from the datasets that are available to train the systems. Neither of these streams of research directly contribute to the evaluation of reliability of AI systems. DA is intended to retrain AI to new input sets, not increase robustness to departures from the training distribution. OOD detection is intended to detect when a departure from the training distribution occurs but says little about how the model will behave or how the system will perform in those cases.

\subsection{OOD Detection is Not Good Enough}
There are some cases in which it is possible to fall back into a safe mode when a system has decreased confidence in its ability to perform safely. In those cases, OOD detection can be directly valuable. That is the case for many of the systems at use today in driver assist technologies where control can be transferred to a human. This, in effect allows systems with low ASIL to be used because Controllability is high. At higher levels of automation\cite{autonomyLevels} that will not be true or will be true to a decreasing degree, so OOD detection becomes less valuable as the degree of automation increases. 

Still, it is valuable to know when a component should be expected to be unreliable. It may be possible to decrease that component's contribution to the system's decisions or actions. For example, an imaging system that suspects it is unreliable might cede control to other complementary components such as lidar or radar. In so doing, the overall system comes to rely on fewer components. The frequency at which that occurs as well as the independence of the various components can limit the safety of the system and the usefulness of that component.

In contrast, many applications of AI and autonomy do not have a natural or acceptably safe option for the system or component that allow it to simply choose not to operate. A car without a steering wheel cannot cede control to a sleeping passenger because the inputs are OOD, it must decide whether to swerve or maintain course regardless of the underlying distribution of the input it receives. The same goes for military systems in a communications-denied environment or perhaps even cyber defense systems that operate too quickly for a human to intervene at a meaningful timescale. For many applications of interest, OOD detection is of some but limited value.

\section{Measuring OOD Performance}
Prior to deployment, the tester or certifier may need to know the degree to which performance is degraded by operating in OOD conditions. The machine itself may also benefit from knowing how much performance is expected to degrade upon encountering an OOD input to evaluate its own confidence.

\label{sec:measuring}
\subsection{OOD Performance Deserves Increased Attention}

There is a small but growing body of work focused on measuring or improving the OOD performance of machine learning models. The sub-field does not seem to have matured to the point of having established consistent terminology so papers refer to the problem in many ways such as OOD robustness,\cite{hendrycksPretraining} OOD generalization,\cite{hendrycksFaces} robustness to distributional shift,\cite{ovadiaDistShift,weihuaHuDistShift} and sometimes the topic is addressed in a paper without using any specific terms or phrases that differentiate in-distribution performance from OOD.\cite{yinFourier,geirhosShapeBias} As a comparison, the "adversarial examples" sub-field is far more unified with respect to terminology and overarching goals despite being in effect a subset of OOD robustness. It is a subset that is focused on a specific context for encountering OOD inputs (an intentional attack) and a confined set of techniques for creating them (imperceptible perturbations, typically using small $l_{p}$-norm). 

Vulnerability to attack is a partially separated issue from certification in a practical sense for many systems. In a more familiar example, draining the brake fluid is a trivial task for an automotive enthusiast but vehicles are not considered brittle as a result. Further, robust designs and testing are not the only means for addressing vulnerability to attack, there are legal and ethical frameworks for addressing them. Still, there are lessons to be learned from the flurry of work.

Creating models that will perform adequately when exposed to adversarial examples is a popular area of research.\cite{dong,cohen,hartnett} Thousands of adversarial examples papers are put out each year and high-profile structured competitions have been devoted to the task of making models that are robust to them.\cite{attackComp2017} Numerous and widely varying methods to improve robustness have been explored and reviewed elsewhere.\cite{advRevChakraborty, advRevQiu} A common approach involves including adversarial examples in the training set, in essence broadening the training distribution so that adversarial examples are no longer OOD. Many approaches focus on compression of the inputs or other input preprocessing steps.\cite{comDefend, certifiedNoise, certifiedSmoothing} Others focus on smoothing the outputs or gradients of the model such as to make neighboring categories farther away in terms of the amount of change needed to the inputs or simply to make the changes harder for attackers to determine.\cite{smoothingInputGradients,smoothingSharpness}

Although these defenses often do not hold up well to changes in adversary behavior,\cite{falseSecurity} some of these approaches are amenable to the broader problem of OOD robustness, especially considering that OOD robustness in general does not usually need to contend with changes in adversary behavior. The snow will not change the way it falls and the sensor in a camera will not change its sensitivity to noise just to interfere with a machine learning algorithm. Drawing from the literature on generalization and robustness to adversarial examples then, smoothed gradients\cite{keskar} are expected to be helpful in improving OOD performance. Additionally, expanding the training data to draw from a wider range of distributions has also been shown to improve OOD robustness, although care is needed to ensure that improving in one type of distributional shift does not detract from others.\cite{geirhosHumans} Hendrycks et al. suggest in a preliminary sense that using larger models, incorporating self-attention, and the use of pretrained models are also means to improve OOD robustness.\cite{hendrycksFaces}

\subsection{AI is Not So Brittle}
\subsubsection{Deep Learning for Image Classification}

Some papers have already shown that the performance of machine learning classifiers fall off as the inputs are degraded in various ways.\cite{Karahan,hendrycksFaces,geirhosHumans} This is not surprising as the information being conveyed through those inputs is degraded. The important question with regard to test and certification is the rate at which performance falls off for various levels of departure from the training distribution. One of the most intuitive machine learning tasks is image recognition and comparison to human performance was attempted in one case by Geirhos et al.\cite{geirhosHumans} In their work, humans outperformed machines on most manipulations but were not always clearly superior. This field is still far from mature so further improvements in model robustness should be anticipated. 

As an illustration of the degree of robustness or fragility of standard machine learning models, we performed ten different manipulations at ten different levels on a thousand randomly selected Image-Net images. The manipulations are listed in table \ref{tab:manipulationsList}.

\begin{table}[!htb]
    \centering
    \begin{tabular}{|c|c|}
        \hline
         \textbf{Manipulation} & \textbf{Maximum Magnitude} \\ \hline
         Gaussian Blur & 10 pixel standard deviation \\ \hline
         Average Blur & 20 pixel window size \\ \hline
         Motion Blur & 25 pixel kernel size \\ \hline
         Gaussian Noise & 150 standard deviation intensity \\ \hline
         Speckle Noise & 1 standard deviation intensity \\ \hline
         Salt and Pepper Noise & 0.3 probability \\ \hline
         Darkening & 220 intensity shift \\ \hline
         Brightening & 220 intensity shift \\ \hline
         Single Occlusion & 150 pixel per side \\ \hline
         Multiple Occlusions & Matched to Single Occlusion coverage \\ \hline
    \end{tabular}
    \caption{Image degradations and their magnitudes}
    \label{tab:manipulationsList}
\end{table}

These altered datasets are OOD from the clean ImageNet dataset on which the neural networks were trained and they are OOD in controlled and measurable ways. The types of departures from the original distribution are both intuitive for humans to see and are likely types of departure that can be expected to occur in real-world scenarios. In most cases, the departures range from slight but noticeable for humans at the lowest levels to substantial but not entirely debilitating for humans at the high end. The speckle and salt and pepper noises were more difficult for the models as compared to humans so they are evaluated over a range that is still substantial but less debilitating for humans.

\begin{figure}[!htb]
  \centering
  \includegraphics[width=1\linewidth]{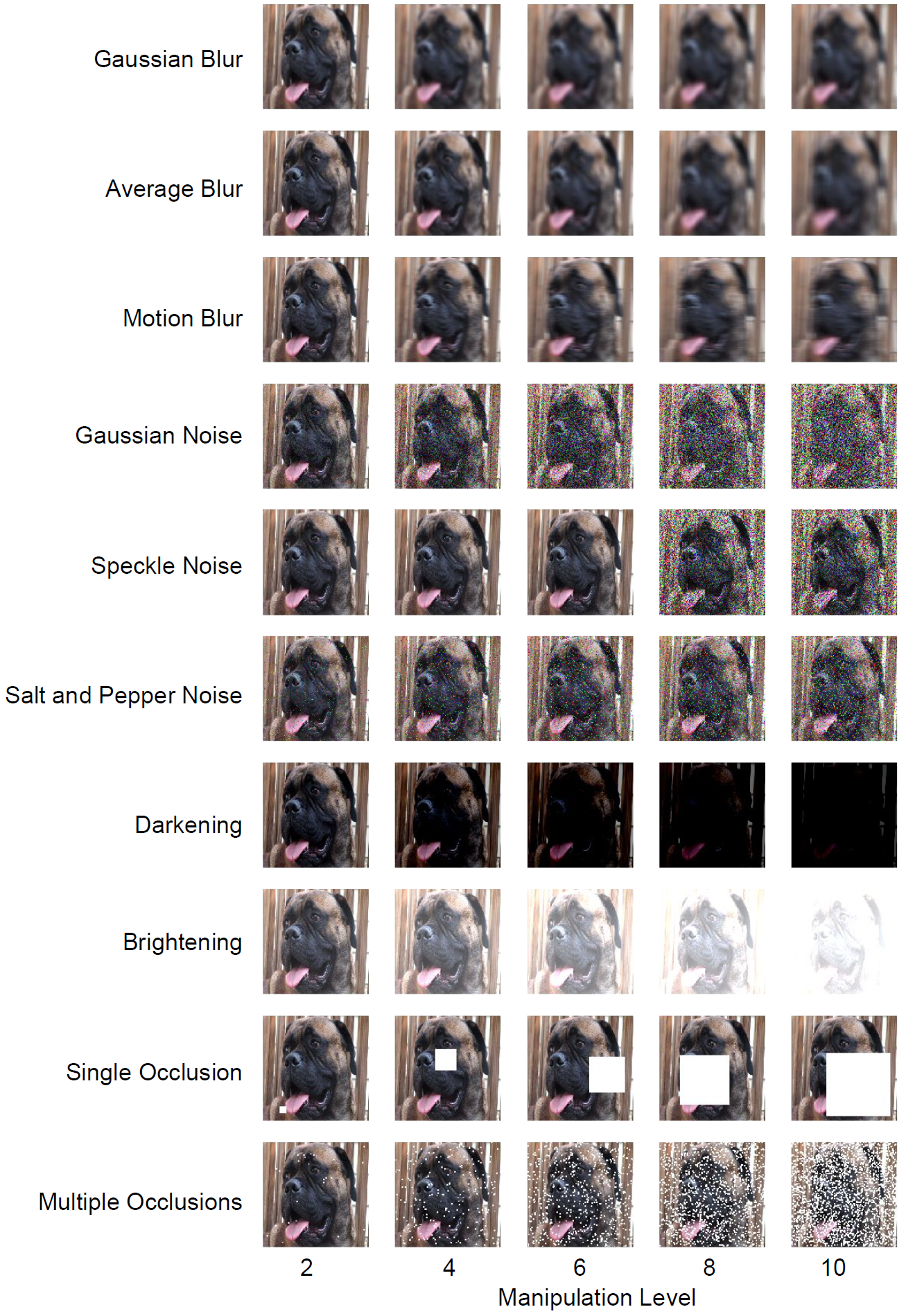}
  \caption{Examples of the degradation mechanisms and their severity.}
  \label{fig:degradationExamples}
\end{figure}

The degree of manipulation is easiest to contextualize in images as shown for one example out of the thousand in the dataset in Figure \ref{fig:degradationExamples}. The dog shown in Figure \ref{fig:degradationExamples} is on the naturally robust end in some ways because it takes up the whole space of the image. If the dog were only contained in a smaller fraction of the image it could be completely masked by an occlusion or, for blurring, more background features would be incorporated into the dog itself.

\begin{figure}[h]
  \centering
  \includegraphics[width=0.9\linewidth]{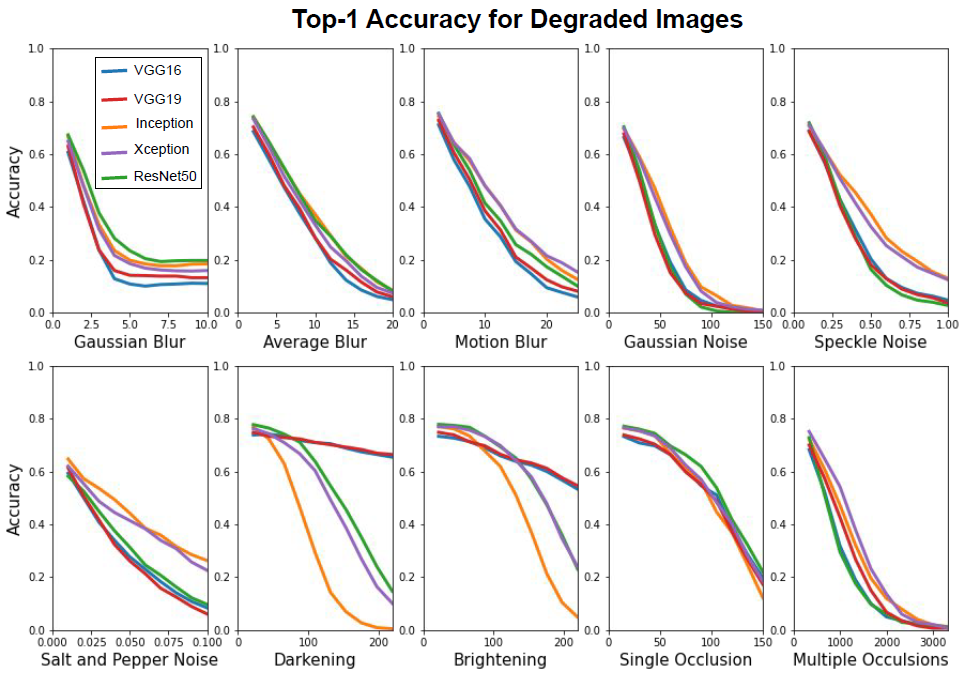}
  \caption{Top-1 degradation is shown for all models and all degradation types.}
  \label{fig:oodPerformance-top1}
\end{figure}

\begin{figure}[h]
  \centering
  \includegraphics[width=0.9\linewidth]{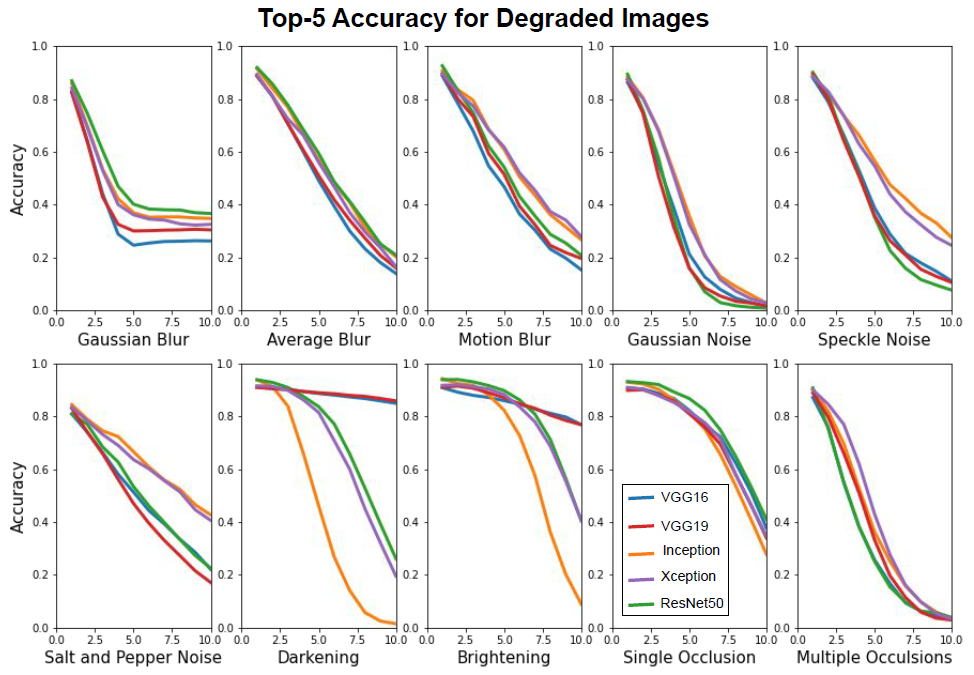}
  \caption{Top-5 accuracy is shown for all models and all degradation types.}
  \label{fig:oodPerformance-top5}
\end{figure}

We then evaluated five neural networks on the altered datasets having made no changes to the networks or their training procedures. The five networks were VGG-16 and VGG-19,\cite{vgg16} InceptionV3,\cite{inception} Xception,\cite{xception} and ResNet-50\cite{resnet50}. The top-1 accuracy for all five models on all ten degradation types is shown in Figure \ref{fig:oodPerformance-top1} and top-5 accuracy is shown in Figure \ref{fig:oodPerformance-top5}.

The qualitative shape of the falloffs in performance are different. In some cases the performance falls off quickly and in others it is more robust. In some it is gradual and in others it is more sudden (remember that speckle and salt and pepper noise are shown over a range that is not often debilitating for humans). The trend observed in the single occlusion case may be somewhat deceptive because the fraction of the image that is occluded grows as the square of the x-axis so it is not surprising to see degradation slow at first then accelerate. It is shown in that way to contrast against the multiple occlusions case which is plotted with the same step size on the x-axis. The number of multiple occlusions was selected to have approximately the same average fraction of the image covered as the single occlusion steps but the two cases show qualitatively different trends. For multiple small occlusions like snow, the performance falls off more linearly at first then saturates. All the models tested were more susceptible to many scattered small occlusions than to one big one even though they block the same fraction of the image.

The Gaussian blur performance fell off approximately linearly then saturated at an accuracy much higher than random guessing. Most of the other degradations appear to trend toward lower accuracies although with noticeably different shapes to the curves. In some cases, most notably the darkening and brightening, different models resulted in not only different overall quantitative performance but with different qualitative trends. VGG16 and VGG19 are comparably stable against brightness and darkness changes. Inception and Xception are comparably stable against noise. And it is not always true that the family of architectures determine the similarity of their fall off in performance. Inception performed poorly on brightening and darkening whereas ResNet-50 and Xception performed more similar to each other.

To determine the relative brittleness of these networks as compared to humans would require a large-scale human evaluation which we do not perform because these networks were not selected for nor designed for OOD performance. The conclusion that OOD performance falls off in a way that is not so starkly different from humans without even being designed for OOD robustness is the intended take-away at this point. Many of the efforts to improve in-distribution generalization, such as adding noise or inducing rotations during training are likely to extend to OOD generalization but that was not the intent in these networks. Alterations to the images in the training set were used but they included actions like image rotations, rescaling, and changes to the color intensity not the distributional shifts evaluate in this paper.\cite{krizhevsky} As models are found or developed to have particularly good OOD performance, they should be baselined against human performance, or more importantly, against the likelihood of encountering similar departures from the training distribution in real-world conditions.

\subsubsection{Deep Learning for Speech Recognition}

Another of AI's contemporary highly-touted successes is in converting spoken language to text. It is common to train and test architectures on one of the available datasets (LibriSpeech, VoxForge, CHiME, etc) as it is for image classification (Image-Net, CIFAR, MNIST, etc). One popular paper though, presenting Deep Speech 2\cite{deepspeech2}, evaluated their trained model using tests drawn from other datasets to determine how performance degrades differently when evaluated on distinct distributions including various sets of standard datasets of text being read, speakers with accents, and noisy datasets.

The various distributions might be subsets of the training distributions. For example the training set may have had Indian accents of the type used in the Indian accent distribution evaluation set. As another example, the simulated noise in the noisy evaluation set may be similar to, or of exactly the same type, as the noise that was added to the training set to improve generalization. Nonetheless, the concept of evaluating performance degradation on particular distributions that are not identical to that of the training examples is clearly on display and is a valuable contribution toward illustrating the robustness or brittleness of AI. 

\begin{table}[!htb]
    \centering
    \begin{tabular}{|c|c|c|c|}
        \cline{2-4}
        \multicolumn{1}{c|}{} & Test Set & AI WER (\%) & Human WER (\%) \\
        \hline
         \multirow{4}{*}{Read} & WSJ eval '92  & 3.10 & 5.03 \\ \cline{2-4}
          & WSJ eval '93 & 4.42 & 8.08 \\ \cline{2-4}
          & LibriSpeech test-clean & 5.15 & 5.83 \\ \cline{2-4}
          & LibriSpeech test-other & 12.73 & 12.69 \\ \hline
         
         \multirow{4}{*}{Accented} & VoxForge American-Canadian  & 7.94 & 4.85 \\ \cline{2-4}
          & VoxForge Commonwealth & 14.85 & 8.15 \\ \cline{2-4}
          & VoxForge European & 18.44 & 12.76 \\ \cline{2-4}
          & VoxForge Indian & 22.89 & 22.15 \\ \hline
          
          \multirow{2}{*}{Noisy} & CHiME eval real  & 21.59 & 11.84 \\ \cline{2-4}
          & CHiME eval sim & 42.55 & 31.33 \\ \hline
    \end{tabular}
    \caption{Word Error Rate (WER) performance of Deep Speech 2 on various distributions of inputs}
    \label{tab:deepspeechResults}
\end{table}

Both Deep Speech 2 and the humans in the experiment experienced degraded performance on some distributions but neither were completely debilitated in any of them. On the easiest distributions (the top four in Table \ref{tab:deepspeechResults}), the AI mostly outperformed the humans. Both had single-digit error percentages as measured by Word Error Rate which is impressive but is again orders of magnitude higher than certifiers are accustomed to for software components in critical systems. 

When the distributions were more specialized (accented and noisy), the humans outperformed the AI. That suggests the AI is less robust than the humans to shifts in distributions but, just as was true for image classification, the fall off is gradual and comparable in magnitude to the humans. Deep Speech 2 was not explicitly designed for OOD performance nor was their experiment designed to evaluate OOD performance, but it is a useful step in that direction.

\subsection{SILs for OOD}
SILs exist for both continuous operation and for low-demand per use operations. Both of those quantities will need to be determined for certain AI applications in critical systems but a third class of SIL may be warranted for AI systems focused OOD performance. For such a metric one would need to establish degrees of OOD and establish acceptable failure rates at those degrees. Even without estimating OOD performance, certifiers will need to estimate the frequency at which OOD inputs will be encountered in the real-world to decide whether the system meets safety thresholds in the real world. If the AI component is the only option or is necessary to help achieve the safety thresholds then OOD detection is insufficient and there is no choice but to estimate the OOD performance.

Estimating the OOD performance will rely on estimating both the frequency and magnitude of departures from the training distribution. There are many means for measuring the divergence between two distributions or the distance from a sample and a distribution. A few common examples are the Mahalanobis and Hausdorff distances and the Kullback-Leibler divergence. The most useful distances though might not be quantitative such as those. It may only be possible for certifiers to envision qualitative or conceptual distances akin to the qualitative levels used in Severity, Exposure, and Controllability that are used to determine the appropriate ASIL in the automotive industry. Qualitative levels such as In-Distribution, Near-Distribution, Somewhat OOD, Far OOD, and Very Far OOD could be useful. The anticipated frequency of encountering situations corresponding to each of those levels could then be matched to required OOD performance at those levels. That OOD performance could be quantitatively measured using real samples or perhaps even supplemented by artificially generated samples from each of the qualitative levels.

\section{Conclusions}
\label{sec:conclusions}

AI is being considered for, or even applied in, critical industries. Those industries have well-established procedures and standards for incorporating new technologies but there are some mismatches between AI and those standards and procedures. First, AI commonly has failure rates that are orders of magnitude higher than the standards used in those critical industries. Second, AI is intended for tasks that do not accommodate straightforward evaluation. The central challenge facing the tester is to estimate the frequency and extent of departures from designed conditions that the AI will encounter in the real world and estimate the AI's performance in those conditions. Fortunately, AI algorithms do maintain some level of performance outside of their design conditions such as on image classification of OOD inputs and perhaps speech recognition with varying accents or noise. If AI can reach the levels of performance in the best of conditions (perfectly in-distribution sampling) that are required for use in critical systems then evaluating performance in non-ideal conditions will be necessary. More work is needed to study the performance degradation of AI algorithms when subjected to OOD samples and to identify ways to improve their OOD performance but it is an area that holds some promise.

\section*{Acknowledgement}
The author would like to thank Open Philanthropy for their interest in this topic as well as Luke Muehlhauser, Gavin Hartnett, Edward Geist, Daniel Ish, Li Ang Zhang, and Peter Whitehead for help collecting and analyzing background materials on existing standards and processes and/or useful conversations about testing for AI and techniques for machine learning in the presence of multiple distributions of data. Naturally, any errors or omission are attributable to the author alone.

\bibliographystyle{unsrt}  
\bibliography{references}

\end{document}